\documentclass[lettersize,journal]{IEEEtran}
\usepackage{amsmath,amsfonts}
\usepackage{amssymb}

\usepackage{bbm}
\usepackage{algorithmic}
\usepackage{algorithm}
\usepackage{array}
\usepackage[caption=false,font=normalsize,labelfont=sf,textfont=sf]{subfig}
\usepackage{textcomp}
\usepackage{stfloats}
\usepackage{url}
\usepackage{orcidlink}
\usepackage{verbatim}

\usepackage{graphicx}
\usepackage{booktabs}
\usepackage{multirow}
\usepackage{colortbl}
\usepackage{color}
\definecolor{citeblue}{RGB}{65,105,225}
\definecolor{tabgray}{RGB}{192,192,192}

\usepackage{cite}

\usepackage{hyperref}
\hypersetup{
	colorlinks=true,
	urlcolor=citeblue,
	linkcolor=citeblue,
	anchorcolor=citeblue,
	citecolor=citeblue
}

\begin{document}

\title{Masked Image Modeling Boosting\\ Semi-Supervised Semantic Segmentation}

\author{Yangyang Li$^{\orcidlink{0000-0002-1328-8889}}$,~\IEEEmembership{Senior Member,~IEEE,}
        Xuanting Hao$^{\orcidlink{0009-0006-1016-5986}}$,
        Ronghua Shang$^{\orcidlink{0000-0001-9124-696X}}$,~\IEEEmembership{Member,~IEEE,}\\
        and Licheng Jiao$^{\orcidlink{0000-0003-3354-9617}}$,~\IEEEmembership{Fellow,~IEEE}

\thanks{
This work was supported in part by XXXXXXXXXXXXXXXXXXXXXXXXXXXX, 
in part by XXXXXXXXXXXXXXXXXXXXXXXXXXXX, 
in part by XXXXXXXXXXXXXXXXXXXXXXXXXXXX, 
and in part by XXXXXXXXXXXXXXXXXXXXXXXXXXXX. 
\textit{(Corresponding author: Xuanting Hao.)}

The authors are with the Key Laboratory of Intelligent Perception and Image Understanding of Ministry of Education, the International Research Center for Intelligent Perception and Computation, the Joint International Research Laboratory of Intelligent Perception and Computation, and the School of Artificial Intelligence, Xidian University, Xi’an, Shaanxi 710071, China (e-mail: 
xt\_hao@stu.xidian.edu.cn).
}

}

\markboth{Journal of \LaTeX\ Class Files,~Vol.~14, No.~8, August~2021}%
{Shell \MakeLowercase{\textit{et al.}}: A Sample Article Using IEEEtran.cls for IEEE Journals}

\IEEEpubid{0000--0000/00\$00.00~\copyright~2021 IEEE}

\maketitle

\begin{abstract}
In view of the fact that semi- and self-supervised learning share a fundamental principle, effectively modeling knowledge from unlabeled data, various semi-supervised semantic segmentation methods have integrated representative self-supervised learning paradigms for further regularization. 
However, the potential of the state-of-the-art generative self-supervised paradigm, masked image modeling, has been scarcely studied. 
This paradigm learns the knowledge through establishing connections between the masked and visible parts of masked image, 
during the pixel reconstruction process. 
By inheriting and extending this insight, we successfully leverage masked image modeling to boost semi-supervised semantic segmentation. 
Specifically, we introduce a novel class-wise masked image modeling that independently reconstructs different image regions according to their respective classes. 
In this way, the mask-induced connections are established within each class, mitigating the semantic confusion that arises from plainly reconstructing images in basic masked image modeling. 
To strengthen these intra-class connections, we further develop a feature aggregation strategy that minimizes the distances between features corresponding to the masked and visible parts within the same class. 
Additionally, in semantic space, we explore the application of masked image modeling to enhance regularization. 
Extensive experiments conducted on well-known benchmarks demonstrate that our approach achieves state-of-the-art performance. 
The code will be available at \url{https://github.com/haoxt/S4MIM}.
\end{abstract}


\section{Introduction}
\IEEEPARstart{S}{emantic} segmentation, a fundamental task in computer vision, involves assigning a category label to each pixel in an image. 
Recently, supervised semantic segmentation using deep neural networks \cite{long2015fully}, \cite{chen2017deeplab}, \cite{zhao2017pyramid}, \cite{ronneberger2015u} has achieved remarkable success. 
Nevertheless, this paradigm demands extensive pixel-wise manual labeling, which is both time-consuming and labor-intensive. 
To alleviate this issue, semi-supervised semantic segmentation \cite{GANfirst} has been proposed. It leverages a small amount of labeled data and a substantial amount of readily accessible unlabeled data for learning. 
In this context, many advanced works \cite{CCT}, \cite{SSCutMix}, \cite{CPS}, \cite{ST}, \cite{ST++} have been developed, in which it is crucial to effectively model knowledge from unlabeled data to further enhance the generalization capability, relying on the semantics acquired from labeled data.

\begin{figure}[!t]
\centering
\includegraphics[width=\columnwidth]{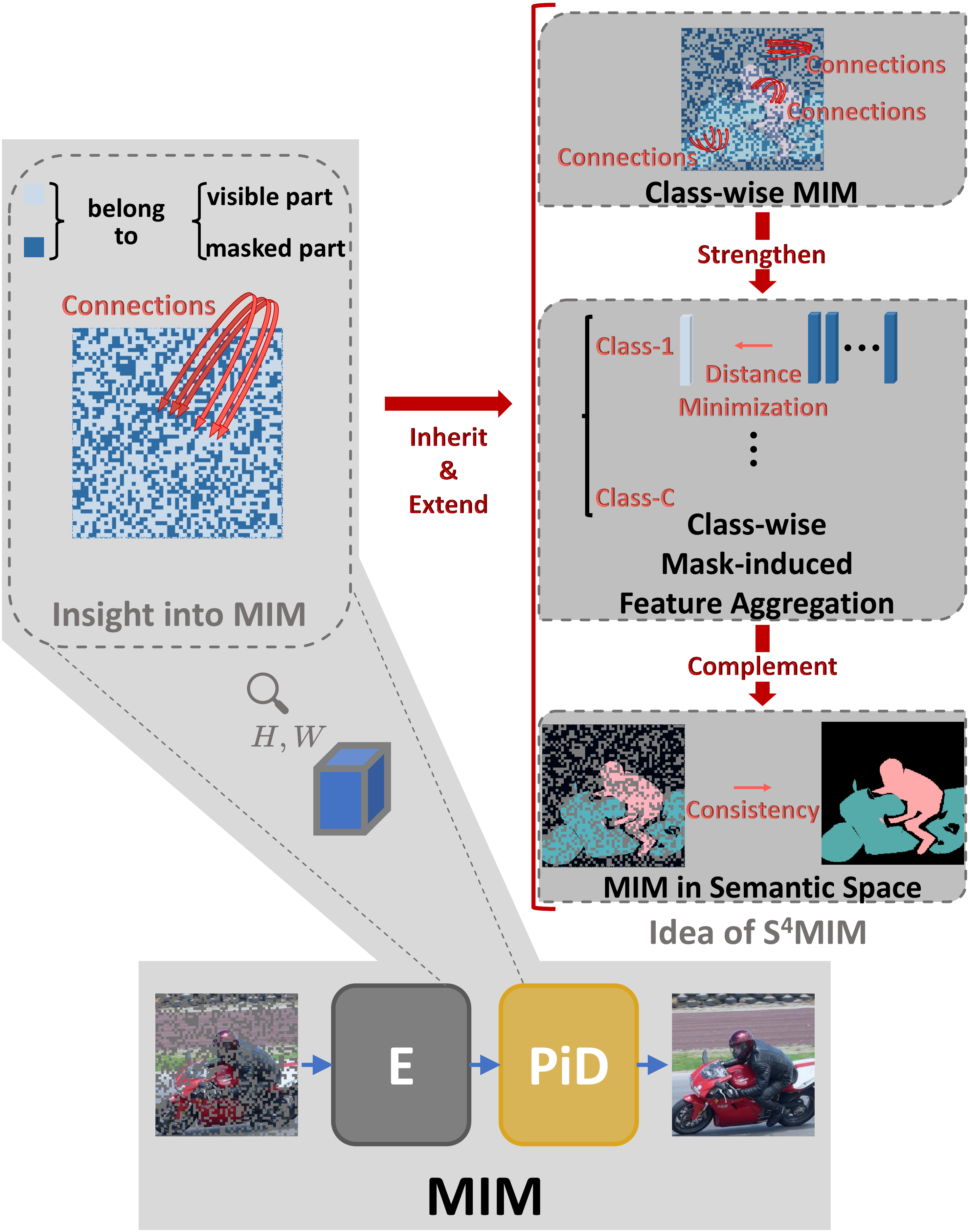}
\caption{\textbf{The illustration of the insight into MIM and the idea of our S\textsuperscript{4}MIM.} 
During the encoding-decoding ($E$-$PiD$) process, MIM learns knowledge through establishing the connections between the features corresponding to the masked and visible parts. 
Building upon this insight, there are three core components of our idea: Class-wise MIM, Class-wise Mask-induced Feature Aggregation, and MIM in Semantic Space. }
\label{fig_idea}
\end{figure}

Concurrently, a paradigm that learns exclusively from unlabeled data, known as self-supervised/unsupervised learning, has received significant attention. 
\IEEEpubidadjcol
Self-supervised learning acquires representation and generalization capabilities by \textit{pretext tasks} defined on unlabeled data, enabling the trained model to serve as a pre-trained model for diverse downstream tasks including classification, semantic segmentation, and object detection. 
According to the difference in pretext tasks, self-supervised learning can be divided into two categories: \textit{discriminative} and \textit{generative}. 
In terms of \textit{discriminative} approaches, typical works include \cite{dis1}, \cite{dis2} and \cite{dis3}, which learn representations by performing pretext tasks such as predicting patch indices, image rotation angles, \textit{etc}. 
In particular, the methods \cite{contrcpc}, \cite{contrmoco}, \cite{contrsimclr}, \cite{contrcluster}, \cite{contrbyol} built upon contrastive learning \cite{contrfirst} have produced promising results, emerging as a prevailing trend in current development. 
On the \textit{generative} side, typical works include \cite{gen1} and \cite{gen2}, which perform pretext tasks such as denoising, colorization, \textit{etc}. 
As a broader extension of denoising, \textit{masked image modeling} (MIM) \cite{masksdae}, \cite{maskcontext} masks a portion of the image, then serves the masked image as input to reconstruct the entire image or the masked content. 
Over the past few years, inspired by the impressive success of masked language modeling in natural language processing (NLP) \cite{maskbert}, further advancements have been achieved \cite{maskbeit}, \cite{masksimmim}, \cite{maskmae}, \cite{maskcnn}. 
Therefore, generative self-supervised learning has regained interest among the research community.

As previously discussed, there is an inherent connection between self-supervised learning and semi-supervised semantic segmentation, \textit{i.e.}, \textbf{effectively modeling knowledge from unlabeled data}. 
Consequently, in the semi-supervised semantic segmentation field, numerous works have integrated the self-supervised learning paradigm to benefit from its merits, where contrastive learning is the primary focus \cite{C1}, \cite{C2}, \cite{PC2Seg}, \cite{ReCo}, \cite{U2PL}. 
\textbf{However, the effectiveness of the state-of-the-art \textit{generative} self-supervised learning paradigm, \textit{MIM}, within semi-supervised semantic segmentation has not been widely investigated.}

Motivated by the success of MIM, this work introduces a novel approach into semi-supervised semantic segmentation. 
As illustrated in Fig. \ref{fig_idea} and investigated in \cite{mimth_nips22} and \cite{mimth_cvpr23}, from the perspective of mask-induced learning, MIM effectively models knowledge through connecting the masked part with the visible (unmasked) part of the masked image during the reconstruction process. 
In order to inherit and extend this insight, our approach consists of a shared encoder followed by parallel pixel and semantic decoders to support explorations in pixel, feature, and semantic spaces, corresponding to ``Class-wise MIM'', ``Class-wise Mask-induced Feature Aggregation'', and ``MIM in Semantic Space'', respectively, which are shown in Fig. \ref{fig_idea}.

Firstly, we observe that in the basic MIM, the connections between the masked and visible parts are derived from the entire image rather than the category objects. 
Consequently, simply introducing the basic MIM, which may connect features from different classes, can lead to confusion. 
To address this issue, we introduce a class-wise modification to the basic MIM, since, in a semi-supervised setting, the availability of limited amounts of labeled data allows category information to be provided via both ground-truth and pseudo-label forms. 
Specifically, in the pixel decoder, through injecting category information, the intermediate features of masked image are grouped by class, where each group’s features have active vectors only at the spatial regions corresponding to that class, with the remaining locations set to zero vectors. 
An independent head is then assigned for each group's features to reconstruct the original pixels of the respective class. 
Finally, in the pixel space, pixels from different classes are combined to form the entire image. 
In such a process, the mask-induced connections are established within each class, mitigating the semantic confusion among classes caused by the basic MIM.

Subsequently, to strengthen the intra-class connections, based on the class-wise grouped features, we explicitly minimize the distances between the features from different parts of the same class. 
Concretely, we maintain a dictionary where entries map to each class. 
Each entry stores a prototype representing its class, which is constructed from the features of the visible part belonging to that class. 
Simultaneously, for each class, the features of the masked part are constrained to move closer to the prototype, resulting in class-wise feature aggregation. 
In the feature space, such a design extends the idea of connecting features from the same class.

On the other hand, to more comprehensively investigate the role of MIM, we also implement it in the semantic space. 
The semantic predictions derived from masked images are ensured to align with those from original images. 
As a complement to our explorations in the pixel and feature spaces, we are able to fully leverage the insight into MIM for regularization.

{\bf{In summary, our main contributions are as follows:}}
\begin{itemize}
\item{
We introduce a novel class-wise MIM into the semi-supervised setting, which independently reconstructs regions corresponding to different category objects of the entire image. 
By inheriting and extending the insight into basic MIM, our class-wise variant establishes intra-class connections and mitigates inter-class confusion.
}
\item{
We develop a class-wise mask-induced feature aggregation strategy, extending the previous point. 
This strategy explicitly minimizes the distances between features of the visible and masked parts within the same class, thereby strengthening the intra-class connections.
}
\item{
We also explore MIM in the semantic space as a complement to the above two points. 
Under masking, our goal is to maintain semantic consistency. 
This exploration allows the insight into MIM to be thoroughly integrated into semi-supervised semantic segmentation.
}
\end{itemize}

Formally, our approach is named \textit{Semi-Supervised Semantic Segmentation with MIM} (S\textsuperscript{4}MIM). 
As the experiments conducted on PASCAL VOC 2012 \cite{data_pascal} and Cityscapes \cite{data_city} demonstrate, our S\textsuperscript{4}MIM achieves state-of-the-art performance. 
The validation of each component's effectiveness proves that MIM can boost semi-supervised semantic segmentation.

\section{Related Works}
\subsection{Semi-supervised Learning}
Semi-supervised learning, initially explored in classification task, has recently been extended to semantic segmentation—an even more complex dense prediction task.

{\bf{Semi-supervised classification.}} 
Two primary strategies have emerged for tackling the classification task: entropy minimization, and consistency regularization. 
Entropy minimization \cite{classst_em} encourages the model to produce highly confident predictions on unlabeled data. In keeping with this strategy, self-training \cite{classst_st}, \cite{classst_stdefense}, \cite{classst_stnoise} assigns pseudo-labels to unlabeled data to provide additional supervision signals. 
Consistency regularization \cite{classcon_pi}, \cite{classcon_mt}, \cite{classcon_vat}, \cite{classcon_ds}, \cite{classcon_mixmatch}, \cite{classcon_remixmatch}, \cite{classcon_uda}, on the other hand, aims to achieve invariant predictions under different perturbations. 
FixMatch \cite{fixmatch}, a concise and effective approach, exploits the advantages of both strategies. It builds consistency between weak and strong perturbations of the same data via pseudo-labels. 
Additionally, S\textsuperscript{4}L \cite{s4l} is the first to introduce self-supervised pretext tasks into semi-supervised classification.

{\bf{Semi-supervised semantic segmentation.}} 
Based on the difference in how to effectively model knowledge from unlabeled data, previous efforts can be generally divided into three categories: approaches based on generative adversarial networks, consistency regularization, and self-training. 
Inspired by generative adversarial networks \cite{GAN}, \cite{GANfirst} and \cite{GAN2} employ the segmentation model as the discriminator or generator to enhance learning. 
Consistency regularization, proven effective in classification, has also led to many outstanding studies \cite{CCT}, \cite{SSCutMix}, \cite{CPS}, \cite{PseudoSeg}, \cite{AEL}, \cite{Atso}, \cite{CCVC}, \cite{AugSeg} in semantic segmentation. 
In the research line of this strategy, CCT \cite{CCT} enhances the contribution of consistency regularization through applying diverse feature-level perturbations. 
French \textit{et al.} \cite{SSCutMix} highlight the important role of CutOut\cite{cutout} and CutMix\cite{cutmix} in the segmentation task. 
CPS \cite{CPS} performs consistency regularization via two separate models learning from each other. 
Meanwhile, self-training \cite{ST}, \cite{ST1}, \cite{ST2}, \cite{ST++} has garnered renewed attention. 
In these works, Yuan \textit{et al.} \cite{ST} propose a DSBN module for utilizing the strong perturbations. ST++ \cite{ST++} further emphasizes the value of appropriate strong perturbations.

UniMatch \cite{UniMatch}, an extension of FixMatch \cite{fixmatch}, designs two powerful techniques, named UniPerb and DusPerb. 
UniPerb involves the application of strong perturbations to data at both image and feature levels. 
Regarding DusPerb, it applies dual image-level strong perturbations. 
These techniques are integrated into FixMatch \cite{fixmatch}, shaping the overall framework. 
In order to investigate our S\textsuperscript{4}MIM under different perturbation levels, while maintaining efficiency, we adopt \textit{FixMatch with UniPerb} \cite{UniMatch} as the semi-supervised baseline.

\subsection{Self-supervised Learning}
Since it shares the core idea with its semi-supervised counterpart, self-supervised learning also concentrates on extracting valuable information from unlabeled data. 
In this field, state-of-the-art methods include discriminative contrastive learning and generative masked image modeling.

{\bf{Contrastive learning.}} 
Contrastive learning \cite{contrfirst} includes three main types of elements: anchors, one positive key and several negative keys for each anchor. 
In the feature space, for each anchor, this paradigm aims at pulling it and the positive key closer, while pushing it and the negative keys farther apart. 
Typically, the anchor and positive key correspond to different augmented versions of the same example, whereas the negative keys correspond to other examples \cite{contrcpc}, \cite{contrmoco}, \cite{contrsimclr}, \cite{contrcluster}. 
In addition, without utilizing negative keys, BYOL \cite{contrbyol} and SimSiam \cite{contrsimsiam} avoid mode collapse by the asymmetric architecture.

When it comes to a semi-supervised setting, category information can be provided, converting contrastive learning from example-wise to \textit{class-wise}. In other words, class-wise contrastive learning constrains features of the same class to be similar, while those of different classes to be dissimilar. 
Leveraging this paradigm, numerous promising methods \cite{C1}, \cite{C2}, \cite{PC2Seg}, \cite{ReCo}, \cite{C3}, \cite{ADCL}, \cite{U2PL}, \cite{DGCL}, \cite{ESL} have been proposed for semi-supervised semantic segmentation.

Within our S\textsuperscript{4}MIM, in the pixel space, MIM is converted to a \textit{class-wise} format, comparable to the transition of contrastive learning from self-supervised to semi-supervised settings. 
Furthermore, in the feature space, we design a \textit{class-wise feature aggregation} strategy, associated with the class-wise contrastive learning method that omits negative keys.

{\bf{Masked image modeling.}} 
The framework of MIM \cite{masksdae}, \cite{maskcontext}, \textit{i.e.}, recovering the entire image or masked area from the masked input, has been proposed for nearly a decade. 
In recent years, the methods based on the random patch masking and transformer architecture, such as SimMIM \cite{masksimmim} and MAE \cite{maskmae}, have significantly advanced, which are driven by the success of BERT \cite{maskbert} in the NLP field. 
Moreover, SparK \cite{maskcnn} introduces the first BERT-style MIM method based on convolutional neural networks (CNNs), overcoming key architectural challenges. 
However, unlike contrastive learning, which is popular in semi-supervised semantic segmentation, MIM requires greater attention.

SegMind \cite{SegMind} incorporates both MIM and contrastive learning in the remote sensing scenario. 
In contrast, our S\textsuperscript{4}MIM centers on the role of mask-induced methods, applying them throughout the entire training process. 
In detail, to adapt to the semi-supervised setting, we introduce a novel \textit{class-wise} masked image reconstruction method. 
Secondly, to extend, we design a \textit{mask-induced feature aggregation} strategy. 
And, we propose a masked image modeling method which \textit{focuses on semantics}, differing from treating masks purely as data augmentation. 
It is also worth noting that our S\textsuperscript{4}MIM lacks a flashy network structure design.

\section{Method}
The goal of our work is to effectively exploit
a labeled image set 
$\mathcal{D}^{L} = \{ (x^{l}_{i}, y^{l}_{i}) \}_{i=1}^{N_{l}}$ 
and
an unlabeled image set 
$\mathcal{D}^{U} = \{ x^{u}_{i} \}_{i=1}^{N_{u}}$ 
to learn a generalized model, 
where in most cases $N_{l} \ll N_{u}$. 
Specifically, $x^{l}_{i} \in \mathbb{R}^{H\times W\times D}$ represents the labeled image with $D$ channels, 
$y^{l}_{i} \in \{0, 1 \}^{H\times W\times C}$ is the corresponding ground truth with $C$ classes, 
and $x^{u}_{i} \in \mathbb{R}^{H\times W\times D}$ denotes the unlabeled image. 
To provide a straightforward yet unambiguous description, in the following, the subscript ``$i$'' is omitted, and the plain ``$\sum$'' represents the summation of batch-size data items.

\subsection{Overview}

\begin{figure*}[!t]
\centering
\includegraphics[width=\textwidth]{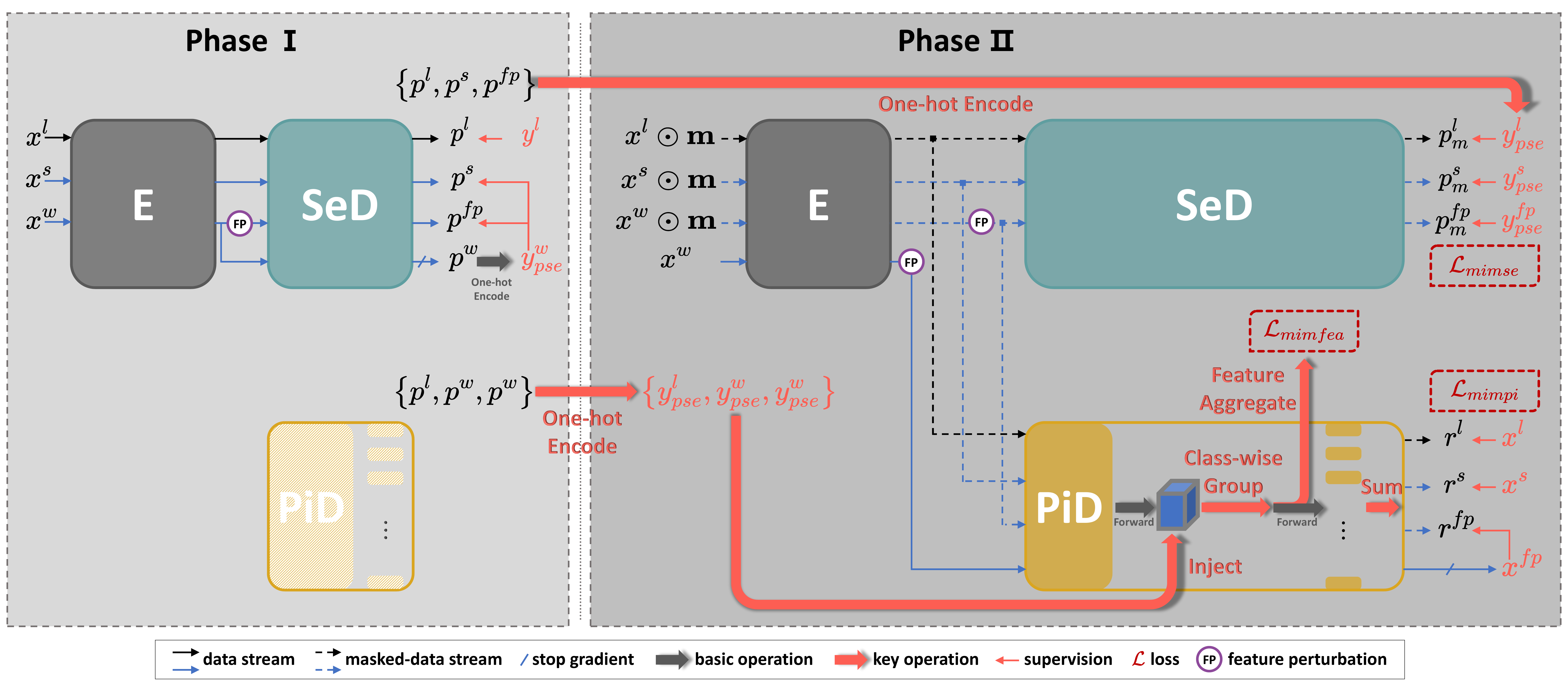}
\caption{\textbf{The overview of our S\textsuperscript{4}MIM.} Each iteration of our approach includes two phases. 
In Phase \uppercase\expandafter{\romannumeral1}, 
as \cite{UniMatch} described, 
labeled data is supervised by ground truth, 
while pseudo-labels generated from weakly perturbed unlabeled data guide its strongly perturbed version. 
In Phase \uppercase\expandafter{\romannumeral2}, 
constrained by pseudo-labels, 
masked data features in the pixel decoder $PiD$ are organized by class-wise grouping. 
These grouped features are then utilized in two branches. 
One branch decodes the grouped features via independent heads, then sums the outputs for final reconstruction. 
The other branch aggregates the features of each group by pulling them closer. 
Meanwhile, at the semantic decoder $SeD$'s output, the masked data is supervised by pseudo-labels derived from the original data.
}
\label{fig_method_total}
\end{figure*}

The overview of S\textsuperscript{4}MIM is illustrated in Fig. \ref{fig_method_total}. 
We design a dual-branch framework in which, following the encoder $E$, 
the semantic decoder $SeD$ targets to predict segmentation classes, 
while the pixel decoder $PiD$ seeks to reconstruct pixel values. 
Depending on whether mask-induced learning is performed, every training iteration is divided into two phases: 
the first corresponds to the semi-supervised baseline, 
and the second to our designed mask-induced learning. 
In each phase, we sample $B_{l}$ labeled images and $B_{u}$ unlabeled images for mini-batch training, where $B_{l}$ is equal to $B_{u}$.

{\bf{Phase \uppercase\expandafter{\romannumeral1}: Semi-supervised baseline.}}
We adopt a strong and efficient approach, FixMatch with UniPerb \cite{UniMatch}, as the baseline in our S\textsuperscript{4}MIM. 
As shown in Fig. \ref{fig_method_total}, during Phase \uppercase\expandafter{\romannumeral1}, 
the prediction $p^{l}$ derived from labeled data $x^{l}$ is supervised by the ground truth $y^{l}$. 
For unlabeled data $x^{u}$, its weakly perturbed version is denoted as $x^{w}$. 
Further perturbations are applied to $x^{w}$ in both the input and feature spaces, resulting in the predictions $p^{s}$ and $p^{fp}$, respectively. 
These predictions, generated under strong perturbations, are supervised by the pseudo-label $y^{w}_{pse}$ from $x^{w}$. 
Accordingly, the losses for labeled and unlabeled data are formulated, respectively, as: 

\begin{equation}
\mathcal {L}_{s} = \frac{1}{B_{l}} \sum 
\ell_{ce}(p^{l}, y^{l}),
\label{eq_s}
\end{equation}

\begin{equation}
\begin{aligned}
\mathcal {L}_{u} = \frac{1}{B_{u}} \sum 
\lambda_{u}
\Psi
\big[ 
&\ell_{ce}(p^{s}, y^{w}_{pse}) + \ell_{ce}(p^{fp}, y^{w}_{pse}) 
\big],
\end{aligned}
\label{eq_u}
\end{equation}

\noindent where $\ell_{ce}$ stands for cross-entropy loss. 
In Eq. \ref{eq_u}, $\lambda_{u}$ represents the trade-off weight, and the indicator function $\Psi = \mathbbm{1}(\max(p^{w})\geq \psi)$ indicates retaining positions in the pseudo-label with high-confidence. 
The $PiD$ stays fixed during this phase.

{\bf{How does MIM learn representations?}} 
As illustrated in Fig. \ref{fig_idea}, masking an image divides it into two components: the visible part and the masked part. 
The model can easily learn the identity mapping for the visible part. However, for the masked part, it must develop a function to transform masks (nothing) into pixel values (something), which is the key point of MIM. 
\textit{Intuitively}, MIM needs to establish connections between the two parts, allowing information from the visible part to be propagated to the masked part for reconstruction. 
\textit{Theoretically}, two recent studies \cite{mimth_nips22}, \cite{mimth_cvpr23} have modeled the MIM framework using different graph models. Nevertheless, both conclude that, under the induction of mask, the model learns high-level representations through the edges that link the vertices of the two parts. 
Given the consistency of conclusions from both intuitive and theoretical perspectives, we build upon this common insight to effectively integrate MIM into semi-supervised semantic segmentation.

{\bf{Phase \uppercase\expandafter{\romannumeral2}: Mask-induced learning.}} 
As presented in Fig. \ref{fig_method_total}, Phase \uppercase\expandafter{\romannumeral2} involves the core idea of our S\textsuperscript{4}MIM. 
Through element-wise multiplication $\odot$, 
the random mask $\mathbf{m} \in \{ \mathbf{0}_{D}, \mathbf{1}_{D} \}^{H\times W}$ 
(with $\mathbf{0}_{D}$ drawn according to the masking ratio) 
is applied to $x^{l}$, $x^{s}$, and $x^{w}$, respectively. 
By leveraging these masked-data streams, we perform mask-induced learning across pixel, feature, and semantic spaces. 
In the $PiD$, pseudo-labels are injected into the features of the masked data. These features are then grouped spatially by class, processed by independent heads, and summed together to reconstruct the entire image in the pixel space, thereby converting the basic MIM into a class-wise variant ($\mathcal {L}_{mimpi}$, Sec. \ref{sec_pix}). 
In the feature space, when further utilizing the class-wise grouped features, we explicitly minimize the distances between features of the visible and masked parts that belong to the same class ($\mathcal {L}_{mimfea}$, Sec. \ref{sec_fea}). 
In addition, in the semantic space, the predictions are constrained to remain semantically consistent despite masking ($\mathcal {L}_{mimse}$, Sec. \ref{sec_sem}).

\subsection{Class-wise MIM in Pixel Space}\label{sec_pix}

In the encoder-decoder ($E$-$PiD$) framework of basic MIM, we redefine the $PiD$ as $Head \circ {PiD}^{\ast}$, where the $Head$ represents the final convolution layer of $PiD$. 
According to this split, the forward process of one masked-data stream is divided into two sub-processes: 

\begin{equation}
fea = {PiD}^{\ast} \circ E(x \odot \mathbf{m}),
\label{eq_mim_forward1}
\end{equation}

\begin{equation}
r = Head(fea).
\label{eq_mim_forward2}
\end{equation}

\noindent The reconstruction result $r$ can be supervised by the original image $x$ with the mean squared error:

\begin{equation}
\ell_{mse}(r, x) = \frac{1}{K} 
\sum_{k=1}^{K} 
{(r\langle k \rangle-x\langle k \rangle)}^{2},
\label{eq_mim_mse}
\end{equation}

\noindent where $K$ is equal to $H \times W \times D$. 
Here, $r\langle k \rangle$ and $x\langle k \rangle$ denote the $k$-th element in the $r$ and $x$, respectively.

\begin{figure}[!t]
\centering
\includegraphics[width=\columnwidth]{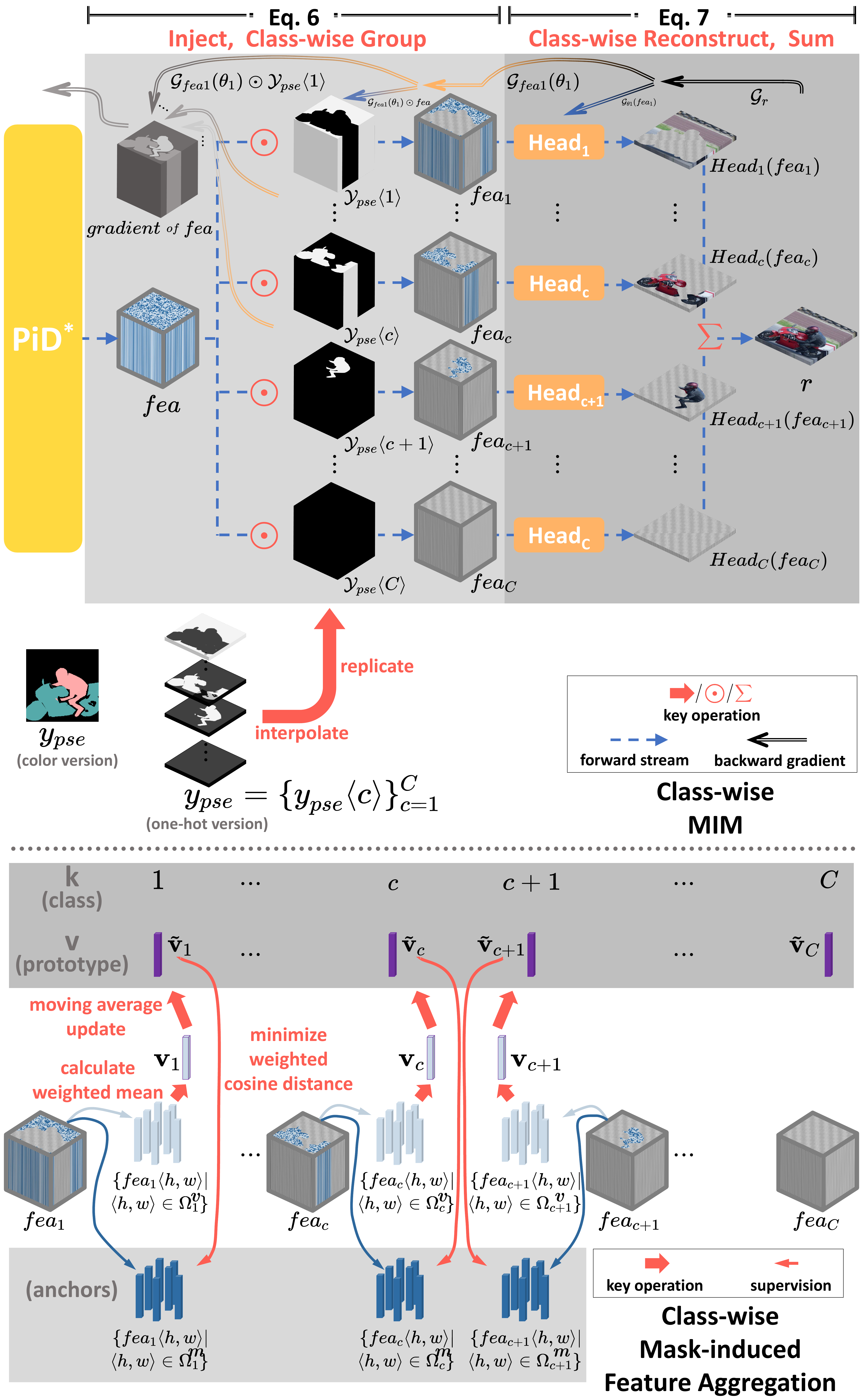}
\caption{
\textbf{The details of our Class-wise MIM (upper area) and Class-wise Mask-induced Feature Aggregation (lower area) with one masked-data stream.} 
In the upper area, $fea$ is multiplied element-wise with each $\mathcal{Y}_{pse}\langle c \rangle$, producing $\{fea_{c}\}_{c=1}^{C}$. 
Each $fea_{c}$ is used with its corresponding $Head_{c}$ to recover the pixels for class $c$. 
By summing these predictions, the entire image is reconstructed. 
In the lower area, within each $fea_{c}$, available spatial positions in the visible and masked parts are denoted as sets $\Omega_{c}^{v}$ and $\Omega_{c}^{m}$, respectively. 
Vectors from $\Omega_{c}^{v}$ are used to construct the class prototype $\widetilde{\mathbf{v}}_{c}$ via weighted mean calculation and moving average updating. 
This prototype supervises the vectors from $\Omega_{c}^{m}$ to aggregate towards it by minimizing the weighted cosine distance.
}
\label{fig_method_pix&fea}
\end{figure}

Revisiting the process represented by Eq. \ref{eq_mim_forward2},  we observe that, in the $fea$, certain spatial positions correspond to the ``$x \odot \mathbf{m}$'' (visible) part, while others correspond to the ``$x \odot (\mathbf{1}-\mathbf{m})$'' (unmasked) part. 
To model knowledge, the $Head$, as a convolution layer, connects these two parts by equally capturing correlations between each position and its surroundings. 
However, in a semi-supervised setting, such a uniform operation disregards semantic information, leading to potential connections between the parts from different classes, which in turn models noisy knowledge.

To address this issue, we aim at \textbf{``establishing connections within each class"} to mitigate semantic confusion. 
As illustrated in the upper area of Fig. \ref{fig_method_pix&fea}, we focus on modifying Eq. \ref{eq_mim_forward2} to convert the basic MIM into a class-wise variant. 
This variant independently reconstructs image regions based on features in $fea$ that correspond to each class. 
First, the pseudo-label $y_{pse}$ generated during Phase \uppercase\expandafter{\romannumeral1} is injected into the $fea$. 
Specifically, in $y_{pse} \in \{0, 1 \}^{H\times W\times C}$, the $c$-th element in the channel dimension (of length $C$) is a binary map indicating whether pixels belong to class $c$, denoted as $y_{pse}\langle c \rangle \in \{0, 1 \}^{H\times W\times 1}$ ($0$ indicates false, $1$ indicates true; $c \in \{1, 2, ..., C\}$). 
Next, to match the dimensions of $fea \in \mathbb{R}^{{H}' \times {W}' \times {D}'}$, the $y_{pse}\langle c \rangle$ is interpolated along the spatial dimensions, and replicated along the channel dimension, referred to as $\mathcal{Y}_{pse}\langle c \rangle$. 
Thus, $fea_{c}$, which contains active vectors only at spatial positions corresponding to class $c$, is defined as:

\begin{equation}
fea_{c} = fea \odot \mathcal{Y}_{pse}\langle c \rangle.
\label{eq_mimpi_feac}
\end{equation}

\noindent In this way, concerning $fea$, the included feature vectors are grouped into $\{fea_{c}\}_{c=1}^{C}$, where $\bigcap_{c=1}^{C} fea_{c}=\varnothing$ and $\bigcup_{c=1}^{C} fea_{c}=fea$. 

After extracting the specific features for each class, we construct a set $\{Head_{c}\}_{c=1}^{C}$, where each element shares the same architecture as the $Head$, but utilizes different initializations. 
Given this set, each $Head_{c}$ is assigned to a corresponding $fea_{c}$ to learn the reconstruction of the current class's pixels.

Finally, all results predicted by $\{Head_{c}\}_{c=1}^{C}$  are combined via a summation operation to reconstruct the entire image. 
Through the above processes, in our Class-wise MIM, the final reconstruction result $r$ can be reformulated from Eq. \ref{eq_mim_forward2} to: 

\begin{equation}
r = \sum_{c=1}^{C} Head_{c}(fea_{c}).
\label{eq_mimpi_r}
\end{equation}

As such, from the perspective of backpropagation, it can be observed that the spatial regions in $fea$ corresponding to different classes exhibit distinct gradient distributions. 
Based on the rules of partial derivative computation and the chain rule of gradient propagation, the analysis of this phenomenon is as follows. 
We notice that our method can be regarded as comprising the summation operation, and the multiplication operation of two variables. 
Given that the gradient received by the current operation node is $\mathcal{G}$. 
For the summation $sum({\{ a_{k} \}_{k=1}^{K}}) = \sum_{k=1}^{K} a_{k}$, the gradient $\mathcal{G}$ is directly propagated to each $a_{k}$. 
For the multiplication $mul(a, b)=a \cdot b$, the gradient with respect to $a$ is $\mathcal{G} \cdot b$, and vice versa. 
Therefore, we employ the ``$\odot$'' in Eq. \ref{eq_mimpi_feac} and the ``$Head_{c}$'' in Eq. \ref{eq_mimpi_r} as checkpoints, 
which equal and include the multiplication operation, respectively. 
In detail, when the gradient propagates, as a variable multiplied by the parameter $\theta_{c}$ of $Head_{c}$, $fea_{c}$ has its gradient $\mathcal{G}_{feac}(\theta_{c})$, which represents a tensor with $\theta_{c}$ as the factor. 
Subsequently, the gradient propagated from $fea_{c}$ to $fea$ is calculated as $\mathcal{G}_{feac}(\theta_{c}) \odot \mathcal{Y}_{pse}\langle c \rangle$. 
This indicates that in $fea$, the gradient of c-th class region $\mathcal{Y}_{pse}\langle c \rangle$ is parameterized by $\theta_{c}$, suggesting the gradients of different classes are determined by their corresponding parameters. 
Afterward, the gradient of $fea$ is propagated to the encoder $E$ to enhance regularization. 
Such an analysis demonstrates that our Class-wise MIM can implicitly establish connections within each class, rather than handling all classes uniformly.

In order to fully exploit data for regularization, as shown in Fig. \ref{fig_method_total}, we utilize three masked-data streams: 
$x^{l} \odot \mathbf{m}$, 
$x^{s} \odot \mathbf{m}$, 
and $x^{w} \odot \mathbf{m}$ with feature perturbation, 
which produce $r^{l}$, $r^{s}$, and $r^{fp}$, respectively. 
As a result, the loss is formulated as: 

\begin{equation}
\begin{aligned}
\mathcal {L}_{mimpi} = \frac{1}{B_{m}} 
\sum 
\lambda_{mp}
\big[ 
&\ell_{mse}(r^{l}, x^{l}) + \\ 
&\ell_{mse}(r^{s}, x^{s}) + \\ 
&\ell_{mse}(r^{fp}, x^{fp})
\big],
\end{aligned}
\label{eq_mimpi}
\end{equation}

\noindent where $\lambda_{mp}$ is the trade-off weight, and $B_{m}$ is equal to $B_{l}$ and $B_{u}$. 
These masked-data streams correspond to the gradient-attached data streams in the Phase \uppercase\expandafter{\romannumeral1}. 
For $x^{l} \odot \mathbf{m}$, due to potential human-defined ignored boundaries in the ground truth $y^{l}$, we utilize the pseudo-label $y^{l}_{pse}$ to guide the class-wise grouping, ensuring that category information covers the entire image. 
For $x^{s} \odot \mathbf{m}$ and $x^{w} \odot \mathbf{m}$, the $y^{w}_{pse}$ is used to provide more accurate category information for these masked-data streams, which are further perturbed separately in the input and feature spaces. 
Notably, unlike $x^{l}$ and $x^{s}$ are readily obtained, $x^{fp}$ is constructed by inputting $x^{w}$ into the model, undergoing the same process as $x^{w} \odot \mathbf{m}$.

\subsection{Class-wise Mask-induced Feature Aggregation}\label{sec_fea}

Subsequently, our goal is \textbf{``to strengthen the connections within each class''} to extend the idea in the previous section. 
In the feature space, considering that the closeness between feature vectors can reflect the strength of their connections, we minimize the distances between vectors from the visible and masked parts within each class. 
However, calculating pairwise distances between vectors from the two parts is computationally intensive. 
To address this, we propose the following strategy, based on the idea of extracting information from the visible part to inform the masked part.

As illustrated in the lower area of Fig. \ref{fig_method_pix&fea}, in each $fea_{c}$, the active vectors are divided into two parts: one corresponding to the visible part, and the other to the masked part. 
During training, due to ignorable boundaries marked in labels and padded areas caused by perturbations, 
we exclude these non-semantic regions from both parts. 
Accordingly, for class $c$, the sets of available spatial positions are denoted as $\Omega_{c}^{v}$ and $\Omega_{c}^{m}$, respectively.

Initially, the vectors corresponding to ``$\Omega_{c}^{v}$'' are consolidated into a single vector, which represents the prototype for the current class. 
In detail, when $\Omega_{c}^{v} \neq \varnothing$, the weighted mean function is adopted to form the prototype $\mathbf{v}_{c}$: 

\begin{equation}
\mathbf{v}_{c} = \frac
{\sum\limits_{\langle h, w \rangle \in \Omega_{c}^{v}}^{} \mathcal{C}onf\langle h, w \rangle \odot fea_{c}\langle h, w \rangle} 
{\sum\limits_{\langle h, w \rangle \in \Omega_{c}^{v}}^{} \mathcal{C}onf\langle h, w \rangle}, 
\label{eq_mimfea_prototype}
\end{equation}

\noindent where $\mathcal{C}onf$ denotes the confidence map of a semantic prediction, derived from spatial interpolation and channel replication. 
The confidence at each position is the maximum predicted value across channels. 
As $fea_{c}$ is a grouped result guided by the pseudo-labels, incorporating a confidence map enables $\mathbf{v}_{c}$ to be formed adaptively, thereby reducing the risk of semantic errors.

Since $\mathbf{v}_{c}$ cannot always be calculated in a batch, we maintain a global dictionary $\{ c : \widetilde{\mathbf{v}}_{c} \}_{c=1}^{C}$ in which the keys are class indices from $1$ to $C$, and the values store the corresponding prototypes across batches. 
Before training begins, each $\widetilde{\mathbf{v}}_{c}$ is initialized to $\mathbf{0}_{{D}'}$. 
In turn, during training, once $\mathbf{v}_{c}$ is obtained in the current batch, the corresponding $\widetilde{\mathbf{v}}_{c}$ is updated through a moving average function: $\widetilde{\mathbf{v}}_{c} = \alpha \widetilde{\mathbf{v}}_{c} + (1-\alpha) \mathbf{v}_{c}$.

In terms of the vectors corresponding to ``$\Omega_{c}^{m}$'', any vector can be denoted as $\mathbf{z}$ without loss of generality. 
The optimization objective for the connection strength between $\mathbf{z}$ and its corresponding prototype is defined as 
one minus the cosine distance, normalized by temperature $\tau$: 

\begin{equation}
\ell_{cos}(\mathbf{z}, \widetilde{\mathbf{v}}_{c}) = 
\Big( 
1 - \frac{\mathbf{z} \cdot \widetilde{\mathbf{v}}_{c}}{\| \mathbf{z} \|_{2} \times \| \widetilde{\mathbf{v}}_{c} \|_{2}} 
\Big) 
/ \tau.
\label{eq_mim_cos}
\end{equation}

\noindent Therefore, the smaller $\ell_{cos}(\mathbf{z}, \widetilde{\mathbf{v}}_{c})$ is, the stronger connection between the two vectors. 
Following the prototype calculation method, 
the loss for all vectors corresponding to $\Omega_{c}^{m}$ is formulated as a confidence-weighted aggregation: 

\begin{equation}
\begin{aligned}
&\ell_{agg}(fea_{c}, conf, \widetilde{\mathbf{v}}_{c}) = \\
&\frac
{\sum\limits_{\langle h, w \rangle \in \Omega_{c}^{m}} 
conf\langle h, w \rangle \times \ell_{cos}(fea_{c}\langle h, w \rangle, \widetilde{\mathbf{v}}_{c})} 
{\sum\limits_{\langle h, w \rangle \in \Omega_{c}^{m}} 
conf\langle h, w \rangle}
,
\end{aligned}
\label{eq_mimfea_agg}
\end{equation}

\noindent where $conf$ refers to the confidence map reshaped to ${H}' \times {W}' \times 1$. 
Benefiting from $\ell_{agg}$, 
the connections between the visible and masked parts within each class can be explicitly strengthened.

As shown in Fig. \ref{fig_method_total}, we also utilize the three masked-data streams: 
$x^{l} \odot \mathbf{m}$, 
$x^{s} \odot \mathbf{m}$, 
and $x^{w} \odot \mathbf{m}$ with feature perturbation to implement our strategy. 
In particular, compared with unlabeled data, the features learned from labeled data can more accurately represent standard semantics, because of the supervision by ground truth. 
Thus, we use the $\{fea_{c}\}_{c=1}^{C}$ of $x^{l} \odot \mathbf{m}$ and the $\mathcal{C}onf$ of $p^{l}$ to update the 
$\{ c : \widetilde{\mathbf{v}}_{c} \}_{c=1}^{C}$. 
Regarding the candidate vectors from the masked parts, 
the $fea_{c}$ and $\Omega_{c}^{m}$ from the three data streams are combined for each class $c$, 
denoted as $\mathbf{fea}_{c}$ and $\mathbf{\Omega}_{c}^{\mathbf{m}}$, respectively. 
In the same manner, the $conf$ of $p^{l}$, $p^{s}$ and $p^{fp}$ are combined as $\mathbf{conf}$ to guide the aggregation process. 
Through this strategy, the aggregation of vectors extends across different data and batches, adhering to a class-wise form. 
The loss is expressed as follows: 

\begin{equation}
\mathcal{L}_{mimfea} = \frac{1}{\sum_{c=1}^{C} \Gamma}
\sum_{c=1}^{C} \lambda_{mf} \Gamma 
\big[
\ell_{agg}(\mathbf{fea}_{c}, \mathbf{conf}, \widetilde{\mathbf{v}}_{c})
\big]
, 
\label{eq_mimfea}
\end{equation}

\noindent where $\lambda_{mf}$ is the trade-off weight, and $\Gamma$ represents the indicator function $\mathbbm{1}(\mathbf{\Omega}_{c}^{\mathbf{m}} \neq \varnothing)$.

Our Class-wise Mask-induced Feature Aggregation is similar to the contrastive learning methods that do not use negative keys \cite{contrbyol}, \cite{contrsimsiam}. 
In our strategy, prototypes serve as positive keys, and available vectors from masked parts act as anchors. 
Within each class, the anchors are aggregated toward the positive key. 
Of note, the primary difference is that, to prevent model collapse into trivial solutions, the methods in \cite{contrbyol} and \cite{contrsimsiam} utilize an asymmetric architecture, while our strategy provides a lower bound by reconstructing diverse pixel values \cite{mimth_nips22}.

\subsection{MIM in Semantic Space}\label{sec_sem}

To fully investigate the role of MIM in semi-supervised semantic segmentation, this section focuses on \textbf{``implementing MIM in a different space''}. 
We adapt the idea of basic MIM, which recovers information at the pixel level, into the semantic space. 
Specifically, to ensure the model can recover the semantics of the masked pixels, 
the cross-entropy loss $\ell_{ce}$ is applied to 
the masked data's predictions and 
the pseudo-labels derived from original data's predictions, both processed via the $E$-$SeD$. 
Importantly, regardless of whether the data is labeled, our method invariably 
performs this semantic consistency learning. 
In this way, the $E$-$SeD$ works toward overcoming the semantic obstacles posed by the masks to facilitate regularization, 
instead of learning standard semantics directly.

As shown in Fig. \ref{fig_method_total}, the same masked-data streams, as previously discussed, are used to produce $p^{l}_{m}$, $p^{s}_{m}$, and $p^{fp}_{m}$, respectively. 
The loss is then formulated as:

\begin{equation}
\begin{aligned}
\mathcal {L}_{mimse} = \frac{1}{B_{m}}  
\sum 
\lambda_{ms}
\big[ 
&\ell_{ce}(p^{l}_{m}, y^{l}_{pse}) + \\ 
&\ell_{ce}(p^{s}_{m}, y^{s}_{pse}) + \\ 
&\ell_{ce}(p^{fp}_{m}, y^{fp}_{pse})
\big],
\end{aligned}
\label{eq_mimse}
\end{equation}

\noindent where $\lambda_{ms}$ represents the trade-off weight.

\subsection{Overall Objective Function}

Our S\textsuperscript{4}MIM is optimized in an end-to-end learning manner. 
The overall objective function is defined as follows:

\begin{equation}
\begin{aligned}
\mathcal {L}_{S\textsuperscript{4}MIM} = 
&(\mathcal {L}_{s} + \mathcal {L}_{u} + 
\mathcal {L}_{mimpi} + \mathcal {L}_{mimfea} + \mathcal {L}_{mimse}) / \\
&(1 + 2\lambda_{u} + 3\lambda_{mp} + 3\lambda_{mf} +3\lambda_{ms}), 
\end{aligned}
\label{eq_mimtotal}
\end{equation}
 
\noindent where $\lambda_{u}$, $\lambda_{mp}$, $\lambda_{mf}$, and $\lambda_{ms}$ are set as 1/2, 1/3, 0.05, and 0.1/3, respectively.

\section{Experiments}

\subsection{Setup}
{\bf{Datasets.}} 
We conduct experiments on two benchmarks. 
PASCAL VOC 2012 \cite{data_pascal} is a dataset with 21 classes. 
It includes 1,464 well annotated images for training and 1,449 images for validation. 
As previous works \cite{CPS}, \cite{U2PL} described, 9,118 coarsely annotated images from the SBD dataset \cite{data_sbd} are used to expand the training set. 
Under such conditions, our S\textsuperscript{4}MIM is evaluated on the \textit{ori} setup (sample labeled data from 1,464 candidates) and the \textit{aug} setup (sample labeled data from 10,582 candidates). 
Cityscapes is a dataset with 19 classes, which faces the urban scene. 
It contains 2,975 fine annotated images for training and 500 images for validation. 

{\bf{Implementation details.}} 
Following previous methods, we adopt ResNet-101 \cite{resnet} pre-trained on ImageNet \cite{data_imagenet} as the encoder and DeepLabv3+ \cite{deeplabv3+} as the semantic decoder. 
The pixel decoder has the same structure as the semantic decoder, differing only by a bias-free 3$\times$3  convolution in the final layer (head). 
Note that, same as UniMatch \cite{UniMatch}, the backbone is set as a speed-up version, and the feature perturbation refers to random channel dropout with a 50\% probability.

For the training of both datasets, we use the stochastic gradient descent (SGD) optimizer, with a poly learning rate schedule for $E$ and $SeD$, while a constant one for $PiD$. 
The batch size is set to 16, comprising equal parts of labeled and unlabeled data. 
For PASCAL VOC 2012, we set input size to 321 (default) or 513, initial learning rate to 0.001, weight decay to 1e$-$4, and training epochs to 80. 
For Cityscapes, we set input size to 801, initial learning rate to 0.005, weight decay to 1e$-$4, and training epochs to 240. 
In particular, following previous works\cite{CPS}, \cite{PS-MT}, \cite{U2PL}, \cite{UniMatch}, the online hard example mining (OHEM) loss is adopted for this dataset training. 
For fair comparisons, in our experiments, both weak and strong perturbations in the input space are implemented exactly as in \cite{UniMatch}. 
Concerning the masking strategy, we mask random patches of the image, with the patch size of 6 and masking ratio of 0.4 by default. 
Besides, $\alpha$ is set to 0.99 to stabilize the update of the class prototype, and $\tau$ is set to 10 to scale the $\ell_{cos}$.

The mean of Intersection over Union (mIoU) is used to evaluate both datasets' validation sets. 
For PASCAL VOC 2012, the mIoU is calculated on the original images.
For Cityscapes, following previous methods, the mIoU is evaluated via a sliding window.

\subsection{Comparison with State-of-the-Art Methods}

\begin{table*}[!t]
\caption{Comparison with SOTAs on the \textbf{\textit{ori} setup of PASCAL VOC 2012}.\\ The number in ``()" denotes the quantity of labeled data.}
\label{tab_pascal_ori}
\centering
\begin{tabular}{l|c c c c c}
\toprule[1.2pt]
\makebox[0.15\textwidth][l]{Method} & 
\makebox[0.13\textwidth][c]{1/16 (92)} & 
\makebox[0.13\textwidth][c]{1/8 (183)} & 
\makebox[0.13\textwidth][c]{1/4 (366)} & 
\makebox[0.13\textwidth][c]{1/2 (732)} & 
\makebox[0.13\textwidth][c]{Full (1464)}\\
\midrule
SupOnly                                & 45.1 & 55.3 & 64.8 & 69.7 & 73.5\\
\midrule
CutMixSeg \cite{SSCutMix}              & 52.16 & 63.47 & 69.46 & 73.73 & 76.54\\
PC\textsuperscript{2}Seg \cite{PC2Seg} & 57.00 & 66.28 & 69.78 & 73.05 & 74.15\\
PseudoSeg \cite{PseudoSeg}             & 57.60 & 65.50 & 69.14 & 72.41 & 73.23\\
CPS \cite{CPS}                         & 64.07 & 67.42 & 71.71 & 75.88 & -\\
ST++ \cite{ST++}                       & 65.2 & 71.0 & 74.6 & 77.3 & 79.1\\
PS-MT \cite{PS-MT}                     & 65.80 & 69.58 & 76.57 & 78.42 & 80.01\\
U\textsuperscript{2}PL \cite{U2PL}     & 67.98 & 69.15 & 73.66 & 76.16 & 79.49\\
FPL \cite{FPL}                         & 69.30 & 71.72 & 75.73 & 78.95 & -\\
CCVC \cite{CCVC}                       & 70.2 & 74.4 & 77.4 & 79.1 & 80.5\\
AugSeg \cite{AugSeg}                   & 71.09 & 75.45 & 78.80 & \textbf{80.33} & \textbf{81.36}\\

UniMatch \cite{UniMatch}               & \underline{75.2} & \underline{77.2} & \underline{78.8} & 79.9 & 81.2\\
\rowcolor{tabgray}
\textbf{S\textsuperscript{4}MIM} & 
\textbf{75.51} & 
\textbf{77.34} & 
\textbf{78.97} & 
\underline{79.98} & 
\underline{81.25}\\
\bottomrule[1.2pt]
\end{tabular}
\end{table*}

\begin{table*}[!t]
\caption{Comparison with SOTAs on the \textbf{\textit{aug} setup of PASCAL VOC 2012}.\\ 
The number in ``()" denotes the quantity of labeled data.
The number after ``\(|\)" denotes the input size of training.}
\label{tab_pascal_aug}
\centering
\begin{tabular}{l|c c c c}
\toprule[1.2pt]
\makebox[0.15\textwidth][l]{Method} & 
\makebox[0.1685\textwidth][c]{1/16 (662)} & 
\makebox[0.1685\textwidth][c]{1/8 (1323)} & 
\makebox[0.1685\textwidth][c]{1/4 (2646)} & 
\makebox[0.1685\textwidth][c]{1/2 (5291)} \\
\midrule
SupOnly \(|\) 321         & 65.6 & 70.4 & 72.8 & 75.4\\
SupOnly \(|\) 513         & 67.5 & 71.1 & 74.2 & 77.2\\
\midrule
CutMixSeg \cite{SSCutMix} & 72.56 & 72.69 & 74.25 & 75.89\\
CPS \cite{CPS}            & 72.18 & 75.83 & 77.55 & 78.64\\
ADCL \cite{ADCL}          & 73.75 & 76.91 & - & -\\
ST++ \cite{ST++}          & 74.7 & 77.9 & 77.9 & -\\
PS-MT \cite{PS-MT}        & 75.50 & 78.20 & 78.72 & 79.76\\
FPL \cite{FPL}            & 74.98 & 77.75 & 78.30 & -\\
CCVC \cite{CCVC}          & 77.2 & 78.4 & 79.0 & -\\
AugSeg \cite{AugSeg}      & 77.01 & 78.20 & 78.82 & -\\
DGCL \cite{DGCL}          & 76.61 & 78.37 & \underline{79.31} & \textbf{80.96}\\

UniMatch \cite{UniMatch} \(|\) 321 & 76.5 & 77.0 & 77.2 & -\\
\rowcolor{tabgray}
\textbf{S\textsuperscript{4}MIM} \(|\) 321 & 
76.82 & 
77.35 & 
77.58 &  
78.41\\

UniMatch \cite{UniMatch} \(|\) 513 & \underline{78.1} & \underline{78.4} & 79.2 & -\\
\rowcolor{tabgray}
\textbf{S\textsuperscript{4}MIM} \(|\) 513 & 
\textbf{78.33} & 
\textbf{78.58} & 
\textbf{79.39} &  
\underline{79.87}\\

\bottomrule[1.2pt]
\end{tabular}
\end{table*}

{\bf{\textit{ori} setup of PASCAL VOC 2012.}} 
As presented in Tab. \ref{tab_pascal_ori}, we compare our S\textsuperscript{4}MIM with other state-of-the-art methods on the \textit{ori} setup of PASCAL VOC 2012. 
Our method shows improvements of 30.41\%, 22.04\%, 14.17\%, 10.28\%, and 7.75\% compared with the supervised only method (SupOnly) across all partitions, respectively. 
In comparison with UniMatch \cite{UniMatch}, our method exhibits a moderate improvement. 
Compared with competing methods, our method achieves optimal results under the 92, 183, and 366 partitions, attaining suboptimal results under the 732 and 1464 partitions. 
Particularly, under the most challenging condition, \textit{i.e.}, the 92 partition, our S\textsuperscript{4}MIM achieves 75.51\% mIoU. 
Moreover, in Tab. \ref{tab_eossbasebone}, we show that our performance improves further when using standard backbone.

{\bf{\textit{aug} setup of PASCAL VOC 2012.}} 
Tab. \ref{tab_pascal_aug} reports the comparative results on the \textit{aug} setup of PASCAL VOC 2012. 
Compared with the SupOnly, our method consistently shows significant performance improvements across different training resolutions. 
In comparison with UniMatch \cite{UniMatch}, our method achieves improvements under various input sizes and partition settings. 
Compared with previous methods, our method gets comparable results with a 513 input size, taking first place under the 662, 1323, and 2646 partitions, and second place under the 5291 partition.

\begin{table}[!t]
\caption{Comparison with SOTAs on the \textbf{Cityscapes}.\\ The number in ``()" denotes the quantity of labeled data.}
\label{tab_city}
\centering
\begin{tabular}{l|c c c c}
\toprule[1.2pt]
\makebox[0.05\textwidth][l]{Method} & 
\makebox[0.065\textwidth][c]{1/16 (186)} & 
\makebox[0.065\textwidth][c]{1/8 (372)} & 
\makebox[0.065\textwidth][c]{1/4 (744)} & 
\makebox[0.065\textwidth][c]{1/2 (1488)} \\
\midrule
SupOnly                            & 66.3  & 72.8  & 75.0  & 78.0\\
\midrule
CCT \cite{CCT}                     & 69.32 & 74.12 & 75.99 & 78.10\\
CPS \cite{CPS}                     & 69.78 & 74.31 & 74.58 & 76.81\\
ADCL \cite{ADCL}                   & 75.82 & 77.20 & 78.76 & -\\
PS-MT \cite{PS-MT}                 & -     & 76.89 & 77.60 & 79.09\\
U\textsuperscript{2}PL \cite{U2PL} & 74.90 & 76.48 & 78.51 & 79.12\\
CCVC \cite{CCVC}                   & 74.9  & 76.4  & 77.3  & -\\
AugSeg \cite{AugSeg}               & 75.22 & 77.82 & \textbf{79.56} & \underline{80.43}\\
ESL \cite{ESL}                     & 75.12 & 77.15 & 78.93 & \textbf{80.46}\\

UniMatch \cite{UniMatch}           & \underline{76.6}  & \underline{77.9}  & 79.2  & 79.5\\
\rowcolor{tabgray}
\textbf{S\textsuperscript{4}MIM} & 
\textbf{76.80} & 
\textbf{78.06} & 
\underline{79.35} &  
79.74\\
\bottomrule[1.2pt]
\end{tabular}
\end{table}


{\bf{Cityscapes.}} 
In Tab. \ref{tab_city}, we show the performance of our method with competitors on the Cityscapes. 
Similarly, this method has advantages compared with UniMatch \cite{UniMatch}. 
From a global perspective, under the 186 and 372 partitions, we achieve the first place. 
Under other partitions, our method attains the second place and the near-second-place performance.

\subsection{Ablation Studies}
Ablation studies are conducted on the \textit{ori} setup of PASCAL VOC 2012 unless otherwise specified.

{\bf{Effectiveness of components.}} 
In our S\textsuperscript{4}MIM, there are three core components: 
Class-wise MIM, 
Class-wise Mask-induced Feature Aggregation, 
and MIM in Semantic Space, 
corresponding to $\mathcal{L}_{mimpi}$, $\mathcal{L}_{mimfea}$, and $\mathcal{L}_{mimse}$, respectively. 
The effectiveness of these components is investigated in Tab. \ref{tab_eoc}. 
To begin with, we adopt FixMatch with UniPerb \cite{UniMatch} as our semi-supervised baseline, achieving 72.26\% mIoU. 
The introduction of $\mathcal{L}_{mimpi}$ significantly outperforms this baseline by 1.99\%. 
This indicates the effectiveness of the proposed Class-wise MIM. 
Building on this, further adding $\mathcal{L}_{mimfea}$ provides a 0.46\% performance improvement, 
revealing that our exploration in the feature space is beneficial for regularization. 
Furthermore, the independent incorporation of $\mathcal{L}_{mimse}$ results in a 1.39\% increase in the baseline.
The combination of $\mathcal{L}_{mimpi}$ and $\mathcal{L}_{mimse}$ yields a 2.48\% improvement. 
Finally, after integrating all the components, our method achieves an improvement of 3.25\% over the baseline, reaching state-of-the-art performance. 
These results highlight the importance of these core components in effectively boosting regularization, with the greatest enhancement observed when all components are utilized.

\begin{table}[!t]
\renewcommand\arraystretch{1.25}
\caption{Ablation study on the \textbf{effectiveness of different components}, including Class-wise MIM ($\mathcal{L}_{mimpi}$), Class-wise Mask-induced Feature Aggregation ($\mathcal{L}_{mimfea}$), and MIM in Semantic Space ($\mathcal{L}_{mimse}$). 
}
\label{tab_eoc}
\centering
\begin{tabular}{c c c | c}
\toprule[1.2pt]
\makebox[0.175\columnwidth][c]{$\mathcal{L}_{mimpi}$} & 
\makebox[0.175\columnwidth][c]{$\mathcal{L}_{mimfea}$} & 
\makebox[0.175\columnwidth][c]{$\mathcal{L}_{mimse}$} & 
\makebox[0.175\columnwidth][c]{92} \\
\midrule
 &  &  & 72.26 \\
$\checkmark$ &  &                          & 74.25 \\

$\checkmark$ & $\checkmark$ &              & 74.71 \\
 &  & $\checkmark$                         & 73.65 \\
$\checkmark$ &  & $\checkmark$             & 74.74 \\
$\checkmark$ & $\checkmark$ & $\checkmark$ & \textbf{75.51} \\
\bottomrule[1.2pt]
\end{tabular}
\end{table}

{\bf{Ablation on ``Class-wise MIM''.}} 
We conduct ablation experiments in Tab. \ref{tab_eomimpi} based on the semi-supervised baseline to demonstrate the necessity of our Class-wise MIM. 
The basic MIM is introduced as a comparison to our method. 
Both methods utilize the same model structure, including $\{Head_{c}\}_{c=1}^{C}$, to prevent the impact of additional parameters. 
It can be observed that introducing the basic MIM provides an improvement compared with the baseline, which indicates the efficacy of MIM framework. 
After we apply the class-wise modification to basic MIM, the performance is further improved, with increases of 1.05\%, 0.48\%, and 0.39\% 
on the 92, 183, and 1464 partitions, respectively. 
This demonstrates the criticality of our class-wise variant, 
which can better handle semantics to adapt to a semi-supervised setting.

\begin{table}[!t]
\renewcommand\arraystretch{1.25}
\caption{Ablation study on \textbf{Class-wise MIM}. 
MIM denotes the basic MIM. Class-wise MIM denotes our method. 
Both methods are based on the same model structure.}
\label{tab_eomimpi}
\centering
\begin{tabular}{c | c c c}
\toprule[1.2pt]
\makebox[0.175\columnwidth][c]{Method} & 
\makebox[0.175\columnwidth][c]{92} & 
\makebox[0.175\columnwidth][c]{183} & 
\makebox[0.175\columnwidth][c]{1464} \\
\midrule
w/o MIM           & 72.26          & 76.01          & 80.14          \\
w/ MIM            & 73.22          & 76.30          & 80.24          \\
w/ Class-wise MIM & \textbf{74.25} & \textbf{76.78} & \textbf{80.63} \\
\bottomrule[1.2pt]
\end{tabular}
\end{table}

{\bf{Ablation on ``Class-wise Mask-induced Feature Aggregation''.}}
In Tab. \ref{tab_eomimfea}, we ablate the confidence \textit{conf} and global dictionary \textit{dic} from our feature aggregation strategy to analyze their roles. 
Under the 92 partition, when neither \textit{conf} nor \textit{dic} is used, the performance decreases by 0.43\% and 0.56\%, respectively. 
This reflects that \textit{conf} and \textit{dic} are introduced to provide better guidance for our strategy. 
Notably, the performance significantly decreases ($\downarrow$1.19\%) when neither \textit{conf} nor \textit{dic} is used concurrently. 
This suggests that implementing our strategy within each batch, guided solely by pseudo-labels, results in insufficient representativeness of class prototypes, which potentially leads to aggregation in the wrong direction. 
Similar observations can be obtained under the 183 partition.

\begin{table}[!t]
\renewcommand\arraystretch{1.25}
\caption{Ablation study on \textbf{Class-wise Mask-induced Feature Aggregation}. 
\textit{dic} denotes maintaining a global dictionary to store the class prototypes. 
\textit{conf} denotes using confidence to calculate the class prototypes and the aggregation loss.
}
\label{tab_eomimfea}
\centering
\begin{tabular}{c | c c c c}
\toprule[1.2pt]

\makebox[0.075\columnwidth][c]{Method} 
& \makebox[0.163\columnwidth][c]{w/o \textit{dic}\&\textit{conf}}
& \makebox[0.163\columnwidth][c]{w/o \textit{dic}}
& \makebox[0.163\columnwidth][c]{w/o \textit{conf}}
& \makebox[0.163\columnwidth][c]{\textit{dic}\&\textit{conf}} \\
\midrule
92  &  74.32 & 74.95 & 75.08 & \textbf{75.51} \\
183 &  76.83 & 77.02 & 77.13 & \textbf{77.34} \\

\bottomrule[1.2pt]
\end{tabular}
\end{table}

{\bf{Ablation on ``MIM in Semantic Space''.}} 
As reported in Tab. \ref{tab_eomimse}, we compare two implementations of MIM in the semantic space, both based on the semi-supervised baseline. 
In the first implementation, we compute the mean squared error loss between the semantic predictions of the masked data and original data. 
The second corresponds to the implementation we adopt. 
Under the 183 and 1464 partitions, the results show that both implementations achieve comparable performance and outperform the baseline. 
However, under the challenging 92 partition, the $\ell_{mse}$ performs worse than the $\ell_{ce}$, showing only marginal improvement over the baseline. 
We believe that in the case where labeled data is very limited, one-hot pseudo-labels can provide clearer semantic guidance than probabilistic predictions.

\begin{table}[!t]
\renewcommand\arraystretch{1.25}
\caption{Ablation study on \textbf{MIM in Semantic Space}. 
$\ell_{mse}$ denotes the mean squared error loss between semantic predictions of the masked and original images. 
$\ell_{ce}$ denotes the cross-entropy loss between semantic predictions of the masked image and pseudo-labels of the original image.}
\label{tab_eomimse}
\centering
\begin{tabular}{c | c c c}
\toprule[1.2pt]
\makebox[0.175\columnwidth][c]{Method} 
& \makebox[0.175\columnwidth][c]{92} 
& \makebox[0.175\columnwidth][c]{183} 
& \makebox[0.175\columnwidth][c]{1464} \\
\midrule
baseline                         & 72.26          & 76.01          & 80.14          \\
w/ $\ell_{mse}$                  & 72.82          & 76.66          & 80.55          \\
w/ $\ell_{ce}$                   & \textbf{73.65} & \textbf{76.67} & \textbf{80.59} \\
\bottomrule[1.2pt]
\end{tabular}
\end{table}

\begin{figure}[!t]
\centering
\includegraphics[width=\columnwidth]{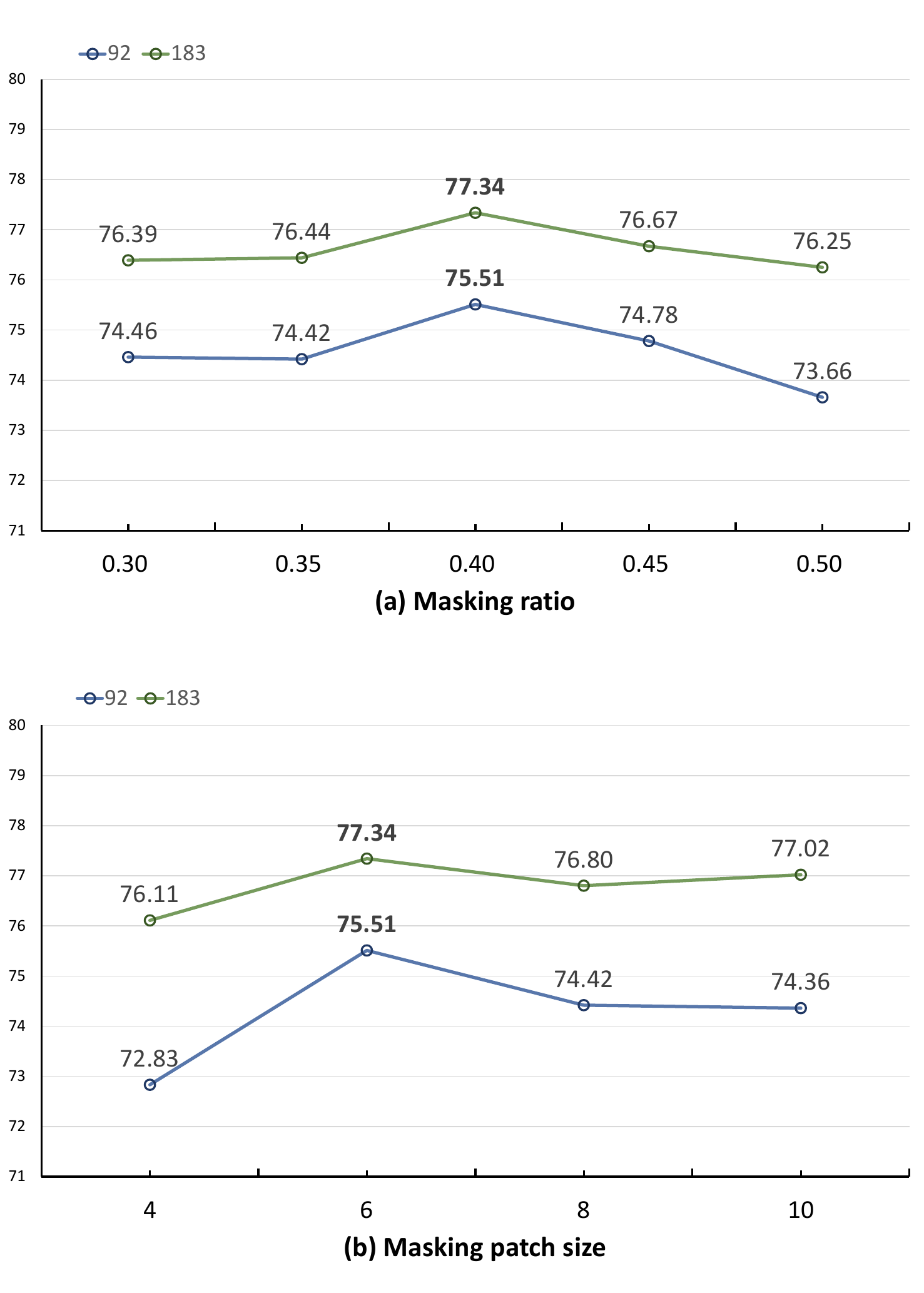}
\caption{Ablation study on the \textbf{parameters of masking}. 
(a) Masking ratio. 
(b) Masking patch size.
}
\label{fig_exp_mrmp}
\end{figure}

{\bf{Ablation on masking ratio and masking patch size.}}
As shown in Fig. \ref{fig_exp_mrmp}(a), we compare the performance of different masking ratios, with a fixed masking patch size of 6. 
It can be observed that the performance remains largely unchanged as the masking ratio varies from 0.30 to 0.35, under the 92 and 183 partitions. 
When the masking ratio reaches 0.40, optimal performance is attained under both partitions. 
Subsequently, an increase in the masking ratio correlates with a decline in the performance. 
At the masking ratio of 0.50, a substantial decline is observed. 
Fig. \ref{fig_exp_mrmp}(b) presents the performance of different masking patch sizes, maintaining a fixed masking ratio of 0.40. 
First, setting the patch size to 4 results in only limited performance. 
Next, increasing the size to 6 achieves the best regularization. 
As the patch size increases to 8 and 10, the performance exhibits a decline followed by a relatively stable state. 
Therefore, we adopt the masking ratio of 0.40 and the masking patch size of 6 as the parameters for our masking strategy.

{\bf{Ablation on backbone.}} 
Furthermore, based on the standard backbone, we show the new performance of our S\textsuperscript{4}MIM in Tab. \ref{tab_eossbasebone}. 
In line with our method, CorrMatch \cite{CorrMatch}, which utilizes the standard backbone and employs FixMatch with UniPerb \cite{UniMatch} as the semi-supervised baseline, is selected for comparison with us. 
We reproduce it using the same experimental setup for both UniMatch \cite{UniMatch} and our method, which consists of 4 GPUs with a batch size of 4 per device. 
It is noteworthy that our method attains additional performance enhancements ($\uparrow$0.47\%, $\uparrow$0.52\%) when using the standard backbone. 
Meanwhile, our method achieves results comparable to those of CorrMatch \cite{CorrMatch}, which represents the latest state-of-the-art. 
These comparisons further validate the efficacy of our S\textsuperscript{4}MIM.

\begin{table}[!t]
\renewcommand\arraystretch{1.25}
\caption{Ablation study on \textbf{Backbone}. \\
``$\dagger$" denotes the method reproduced by us.
}
\label{tab_eossbasebone}
\centering
\begin{tabular}{c | c c}
\toprule[1.2pt]
\makebox[0.10\columnwidth][c]{Method} 
& \makebox[0.175\columnwidth][c]{92} 
& \makebox[0.175\columnwidth][c]{183} \\
\midrule
S\textsuperscript{4}MIM                                 & 75.51          & 77.34     \\
\midrule
CorrMatch\textsuperscript{$\dagger$} \cite{CorrMatch}      & 75.87          &  77.82         \\
S\textsuperscript{4}MIM w/ standard backbone              & \textbf{75.98} & \textbf{77.86} \\
\bottomrule[1.2pt]
\end{tabular}
\end{table}

\subsection{Qualitative results}
\begin{figure}[!t]
\centering
\includegraphics[width=\columnwidth]{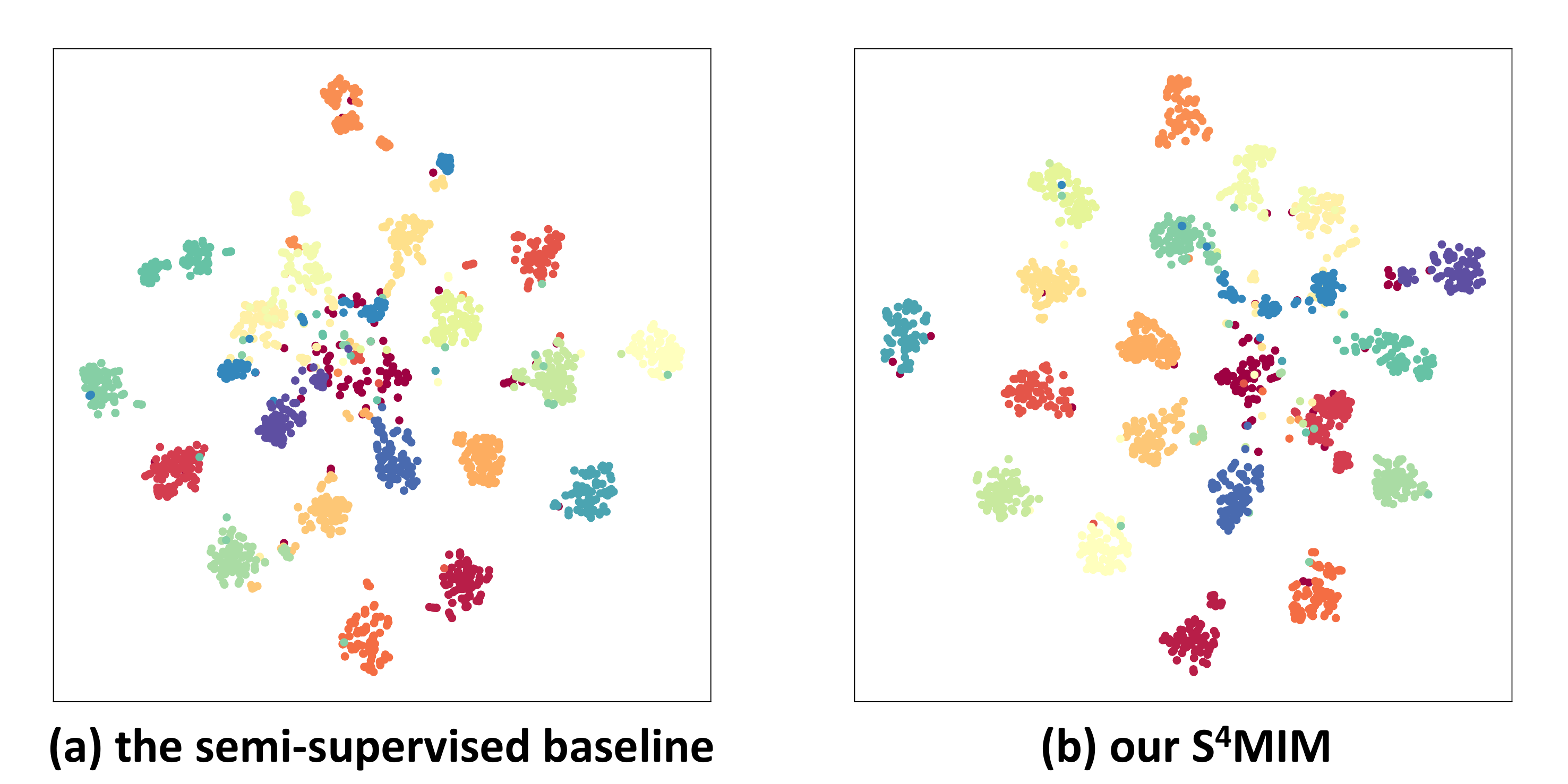}
\caption{Visualization of \textbf{the feature space in $\mathbf{SeD}$} on PASCAL VOC 2012 under 92 partition, using t-SNE \cite{tsne}. 
(a) The semi-supervised baseline. (b) Our S\textsuperscript{4}MIM.
}
\label{fig_exp_tsne}
\end{figure}

Fig. \ref{fig_exp_tsne} depicts the feature space in $SeD$, learned by the semi-supervised baseline and our S\textsuperscript{4}MIM, respectively. 
It is noted that our approach effectively separates several classes that are confused in the baseline, 
which indicates its capacity to boost regularization.

\begin{figure}[!t]
\centering
\includegraphics[width=\columnwidth]{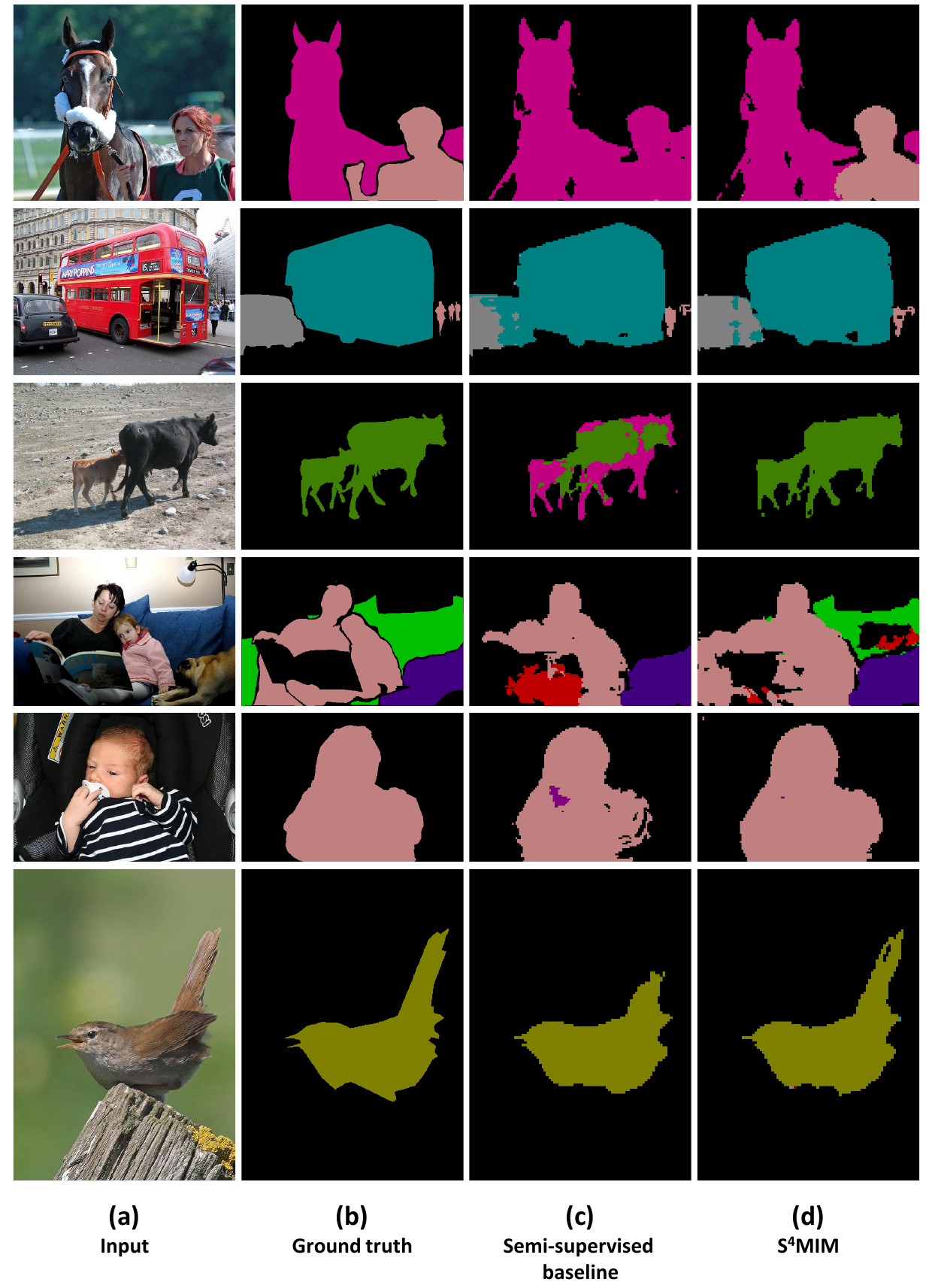}
\caption{Qualitative results on \textbf{PASCAL VOC 2012} under 92 partition. 
(a) Input images. (b) Ground truth. 
(c) The semi-supervised baseline. (d) Our S\textsuperscript{4}MIM.
}
\label{fig_exp_quapas}
\end{figure}

\begin{figure}[!t]
\centering
\includegraphics[width=\columnwidth]{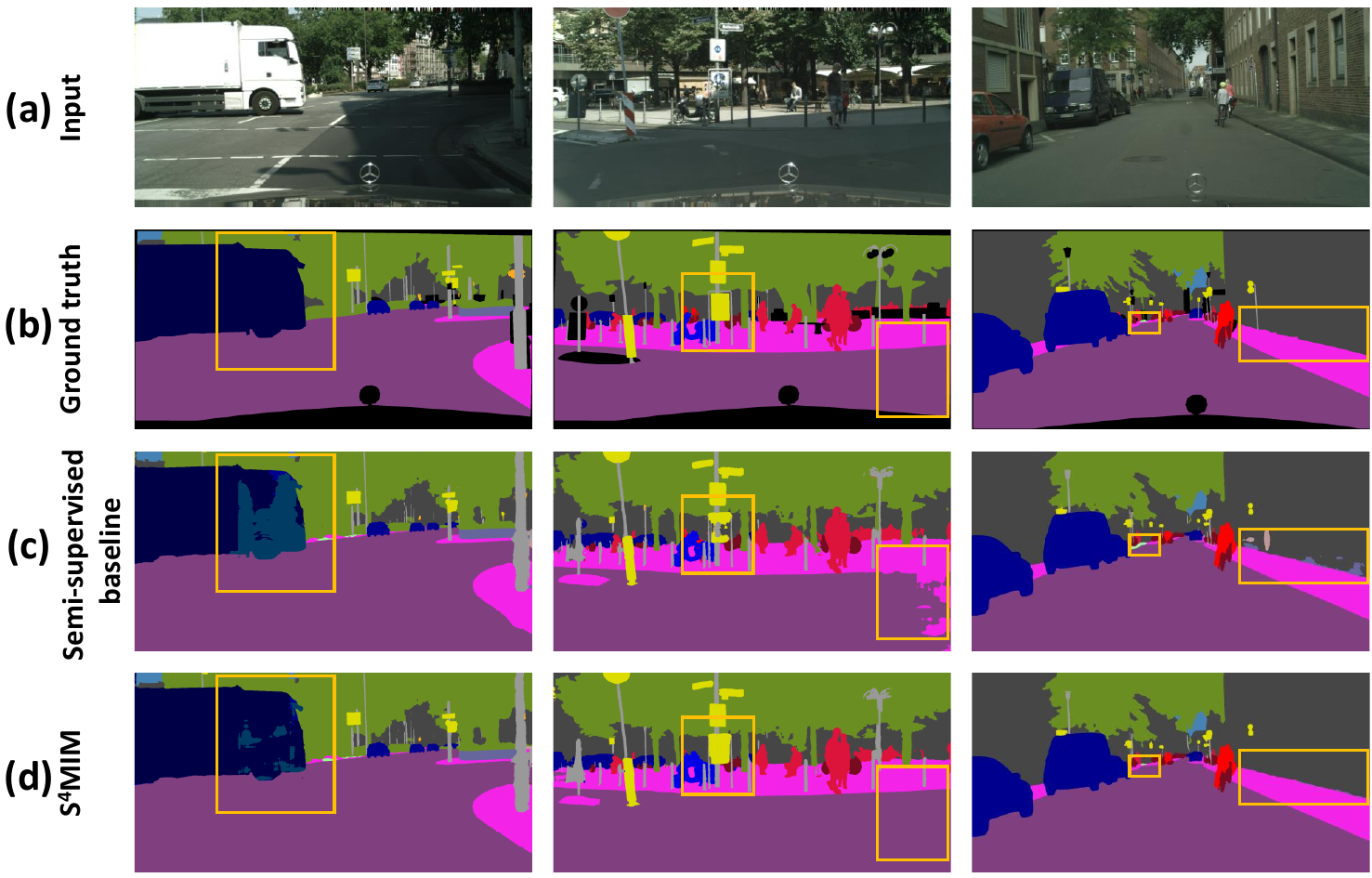}
\caption{Qualitative results on \textbf{Cityscapes} under 744 partition. 
(a) Input images. (b) Ground truth. 
(c) The semi-supervised baseline. (d) Our S\textsuperscript{4}MIM.
}
\label{fig_exp_quacity}
\end{figure}

Fig. \ref{fig_exp_quapas} presents the segmentation results on PASCAL VOC 2012. 
From the first two rows, it can be observed that our method distinguishes adjacent objects more effectively than the semi-supervised baseline. 
In the middle two rows, our method identifies more accurate semantics. 
Additionally, in the last two rows, our method shows a stronger ability to capture the object boundaries. 
Regarding Cityscapes, as shown in Fig. \ref{fig_exp_quacity}, our method provides improved handling of details. 
These observations point out the effectiveness of the mask-induced learning we propose.

\section{Conclusions}

We notice the potential of MIM, the state-of-the-art generative self-supervised learning paradigm, in semi-supervised semantic segmentation. 
Inspired by the insight into MIM, we propose a S\textsuperscript{4}MIM framework to utilize mask-induced learning for regularization. 
First, to establish mask-induced connections within each class, we introduce a novel class-wise MIM to adapt to semi-supervised setting. 
To strengthen these intra-class connections, we develop a class-wise mask-induced feature aggregation strategy aimed at reducing the distances between the features of the masked and visible parts within the same class. 
Furthermore, we implement MIM in the semantic space to fully leverage its inherent function. 
By equipping these components, S\textsuperscript{4}MIM achieves significant improvement compared with the semi-supervised baseline, and attains optimal results across multiple partitions. 
Finally, we hope that this work will encourage the semi-supervised semantic segmentation research community to focus on MIM and integrate it into their studies.

\bibliographystyle{IEEEtranbib}
\bibliography{./main_arxiv}

\vspace{11pt}

\begin{IEEEbiographynophoto}{Yangyang Li}
(Senior Member, IEEE) received the B.S. and M.S. degrees in computer science and technology and the Ph.D. degree in pattern recognition and intelligent system from Xidian University, Xi’an, China, in 2001, 2004, and 2007, respectively. 

She is currently a Professor with the School of Artificial Intelligence, Xidian University. 
Her research interests include quantum machine learning, swarm intelligence, and computer vision.
\end{IEEEbiographynophoto}

\begin{IEEEbiographynophoto}{Xuanting Hao}
received the M.S. degree in computer science and technology from Xidian University, Xi’an, China, in 2022, where he is currently pursuing the Ph.D. degree with the Key Laboratory of Intelligent Perception and Image Understanding of the Ministry of Education of China, School of Artificial Intelligence.

His research interests include deep learning, semi-supervised learning, and domain adaptation.
\end{IEEEbiographynophoto}

\begin{IEEEbiographynophoto}{Ronghua Shang}
(Member, IEEE) received the B.S. degree in information and computation science and the Ph.D. degree in pattern recognition and intelligent systems from Xidian University, Xi’an, China, in 2003 and 2008, respectively.

She is currently a Professor with Xidian University. Her research interests include evolutionary computation, image processing, and data mining.
\end{IEEEbiographynophoto}

\newpage

\begin{IEEEbiographynophoto}{Licheng Jiao}
(Fellow, IEEE) received the B.S. degree from Shanghai Jiao Tong University, Shanghai, China, in 1982, and the M.S. and Ph.D. degrees from Xi’an Jiaotong University, Xi’an, China, in 1984 and 1990, respectively, all in electronic engineering.

From 1990 to 1991, he was a Post-Doctoral Fellow with the National Key Laboratory for Radar Signal Processing, Xidian University, Xi’an. Since 1992, he has been a Professor with the School of Electronic Engineering, Xidian University. He is currently the Director of the Key Laboratory of Intelligent Perception and Image Understanding of Ministry of Education of China, Xidian University. He is in charge of about 40 important scientific research projects and published more than 20 monographs and 100 papers in international journals and conferences. His research interests include image processing, natural computation, machine learning, and intelligent information processing.

Dr. Jiao is also a member of the IEEE Xi’an Section Execution Committee and the Chairperson of the Awards and Recognition Committee, the Vice Board Chairperson of the Chinese Association of Artificial Intelligence, the Councilor of the Chinese Institute of Electronics, a Committee Member of the Chinese Committee of Neural Networks, and an Expert of Academic Degrees’ Committee of the State Council.
\end{IEEEbiographynophoto}

\vfill

\end{document}